\documentclass[]{interact}

\usepackage{epstopdf}
\usepackage[caption=false]{subfig}
\usepackage[T1]{fontenc}
\usepackage{amssymb}
\usepackage{amsmath}
\usepackage{graphicx}
\usepackage{subfig}
\usepackage{comment}
\usepackage[misc]{ifsym}
\usepackage{float}
\usepackage{booktabs}
\usepackage{multirow}
\usepackage{url}
\usepackage[dvipsnames]{xcolor}
\usepackage{footnote}
\usepackage{stfloats}
\makesavenoteenv{tabular}
\makesavenoteenv{table}

\begin{document}


%
\title{Know your sensORs --- A Modality Study\\For Surgical Action Classification}
%
%
%
%

\author{                                                                                                                                          
  \name{Lennart Bastian\textsuperscript{1}\textsuperscript{(\Letter)}\thanks{\Letter: lennart.bastian@tum.de}, Tobias Czempiel\textsuperscript{1}, Christian Heiliger\textsuperscript{2},
  Konrad Karcz\textsuperscript{2},\\ Ulrich Eck\textsuperscript{1}, Benjamin Busam\textsuperscript{1}, Nassir Navab\textsuperscript{1,3}
  }          
  \affil{
    \textsuperscript{1}Chair for Computer Aided Medical Procedures, TU Munich, Germany; \\
    \textsuperscript{2}Minimally Invasive Surgery, University Hospital of Munich (LMU), Germany; \\
    \textsuperscript{3}Computer Aided Medical Procedures, John Hopkins University, Baltimore, USA;
  }
}

\maketitle
\begin{abstract}
The surgical operating room (OR) presents many opportunities for automation and optimization. Videos from various sources in the OR are becoming increasingly available. The medical community seeks to leverage this wealth of data to develop automated methods to advance interventional care, lower costs, and improve overall patient outcomes. Existing datasets from OR room cameras are thus far limited in size or modalities acquired, leaving it unclear which sensor modalities are best suited for tasks such as recognizing surgical action from videos. This study demonstrates that the task of surgical workflow classification is highly dependent on the sensor modalities used. We perform a systematic analysis on several commonly available sensor modalities, evaluating two commonly used fusion approaches that can improve classification performance. Our findings are consistent across model architectures as well as separate camera views. The analyses are carried out on a set of multi-view RGB-D video recordings of 16 laparoscopic interventions.

\end{abstract}

\section{Introduction}




Digitization of the surgical operating room (OR) has long been sought by the scientific and medical communities~\cite{maier-heinSurgicalDataScience2021}.
The analysis of surgical videos is no longer limited to medical devices such as endoscopic cameras -- in the past years, several works have explored the use of ceiling-mounted cameras in an effort to understand OR workflows from an outside perspective. 
As the amount of data stemming from OR sensors increases, new questions arise, such as how to best integrate various modalities into automated surgical systems~\cite{huaulmePEgTRAnsferWorkflow2022} or where to optimally place cameras for specific tasks~\cite{hanelEfficientGlobalOptimization2021,liRobotic3DPerception2020}.
This study seeks to understand which camera modalities are best suited for surgical action recognition, exploring their relative performance in a unique set of multi-view surgical recordings.

Classifying surgical workflow procedures from external OR cameras has been identified as a critical task in developing context-aware systems (CAS) for the operating room ~\cite{schmidtMultiviewSurgicalVideo2021,sharghiAutomaticOperatingRoom2020,twinanda2016multi}.
Temporally aggregating features extracted from video frames enables an understanding of complex workflow tasks that single image analysis alone could not achieve. The deployment of such systems holds promise for many applications, such as optimizing
real-time OR scheduling, designing context-aware intelligent systems, and enabling autonomous anomaly detection \cite{maier2018surgical,maier-heinSurgicalDataScience2021}. However, existing datasets have thus far been limited in size~\cite{srivastavMVORMultiviewRGBD2021,twinanda2016multi}, or modalities captured~\cite{sharghiAutomaticOperatingRoom2020}, leaving many questions on how to build automated systems for the OR unanswered. 

Modern commercial sensors are typically able to acquire RGB, depth, and infrared video streams, although regulatory and technological constraints continue to make acquisition, storage and access challenging in clinical settings~\cite{maier-heinSurgicalDataScience2021}.
While we can glean substantial knowledge from the computer vision community on integrating these modalities optimally~\cite{jung2021wild,lopez2020project}, the OR presents particular challenges. Inconsistent lighting conditions, homogeneous color scales, and ubiquitous occlusions can yield unexpected results when adapting state-of-the-art methods from other domains.
RGB-based algorithms may demonstrate superior performance for distinguishing objects with significant color differences, while depth images are likely better suited for classification tasks under occlusions or poor lighting conditions due to the inherent 3D geometry they represent.
Understanding which modalities contribute most to specific machine learning tasks is crucial for developing high-performing intelligent systems in such a complex environment.

\begin{figure}[tbp]
  \centering
  \includegraphics[width=\textwidth]{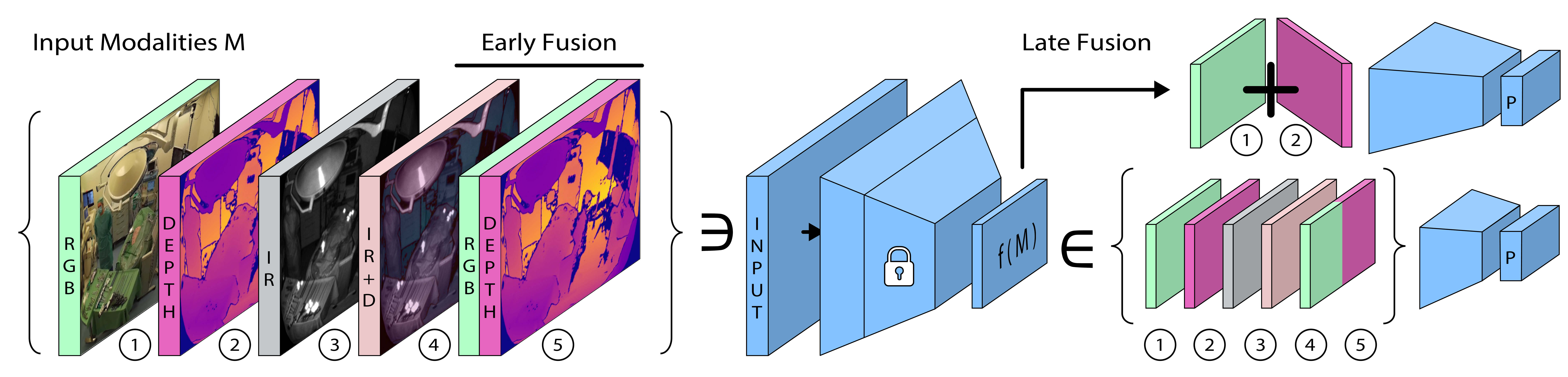}
  \caption{\textbf{Multimodal Data Study for Surgical Workflow Classification}. Different Input Modalities $M$ and their direct combination (left) are fed into a pretrained feature extractor $f$ with partly frozen weights (centre) to extract a feature vector $f(M)$. Depending on the input (1) RGB, (2) Depth, (3) Infrared, (4) Infrared with overlayed Depth, (5) Concatenated RGB+D, different feature extractors are trained. The extracted features (1) to (5) are either directly used (lower right) to predict the surgical phase (P) or combined in a late fusion manner (upper right).}
  \label{fig:pipeline}
\end{figure}

This work seeks to understand which modalities are best suited for surgical workflow recognition in the OR. 
We opt for simple architectures to methodically evaluate action classification performance, making better use of the wealth of data frequently captured in surgical environments.
In summary, our contributions are twofold:
\begin{itemize}
    \item We perform a unique multi-modal analysis for the task of surgical workflow classification, comparing the influence of various modalities (RGB, Depth, IR) on performance. Early and late sensor fusion are investigated together with architectural design choices (see Fig.~\ref{fig:pipeline}).
    \item We demonstrate that modality fusion strategies can be adapted across both modern frame and clip based feature extraction backbones, and perform consistently across different camera positions. We conduct a comprehensive evaluation of our methods on a series of multi-view laparoscopic interventions.
\end{itemize}


\section{Related Works}

\textbf{Surgical Phase Recognition.} 

The past decade has seen the field of surgical phase recognition evolve rapidly, particularly with the advent of machine learning. The first generation of context-aware systems provided surgeons with personalized data before a surgical procedure~\cite{lalys2014surgical}. 
Generally, early data-driven methods focused on classifying phases through data such as information about the presence and absence of tools, the state of the patient, or from specific low-level surgical activities (i.e., cut, swab, drill, sew, etc.)~\cite{forestier2015automatic}. 
Random-forest or SVM-based classifiers were used to analyze various signals from surgical instruments~\cite{blum2010modeling}, for instance, classifying intra-operative activities during laparoscopic cholecystectomies~\cite{stauder2014random}. 
The earliest video-based surgical phase recognition methods used SIFT descriptors and an SVM to classify surgical gestures~\cite{bejar2012surgical}. However, the performance of hand-crafted feature extractors from surgical images has been significantly improved upon by deep learning-based methods.
Furthermore, modern neural network architectures can take long-term temporal context into account, whereas such an analysis was previously only feasible for shorter action segments~\cite{twinandaEndoNetDeepArchitecture2016}.

Video action recognition has seen a surge in attention over the past several years, particularly with the rise of large-scale video datasets such as the Kinetics Human Action Video dataset~\cite{carreiraShortNoteKinetics6002018} or HowTo100M~\cite{miechHowTo100MLearningTextVideo2019}.
These datasets have inspired various novel neural network architectures~\cite{carreiraQuoVadisAction2017,bertasiusSpaceTimeAttentionAll2021} such as X3D~\cite{feichtenhoferX3DExpandingArchitectures2020}, tailored to aggregate spatio-temporal features in RGB videos.
In these settings, action recognition is typically performed on short image sequences, referred to as "clips", which average about 10 seconds in length for Kinetics~\cite{kayKineticsHumanAction2017}.

Surgical workflow analysis from external cameras has been limited to a few internal datasets ~\cite{sharghiAutomaticOperatingRoom2020,srivastavMVORMultiviewRGBD2021,twinanda2015data,twinanda2016multi}.
Despite publishing both RGB and depth sequences, the MVOR~\cite{srivastavMVORMultiviewRGBD2021} dataset consists of 732 frames at relatively low FPS, making video action recognition infeasible. 
A dataset to understand how surgical action recognition performance varies under different sensor input modalities is not currently publicly available. 
While~\cite{twinanda2016multi} explore late fusion for surgical action recognition on single image frames, they only explore a single fusion strategy, and do not consider how clip-based architectures or different camera views could exacerbate these differences.
We therefore acquired a series of 18 OR videos with RGB, infrared, and depth, to better understand these potential differences.

The methodologies used for surgical workflow analysis from videos draw similarity from surgical phase recognition of laparoscopic and endoscopic videos, which have been well established in the
community~\cite{twinandaEndoNetDeepArchitecture2016,czempielTeCNOSurgicalPhase2020,garrowMachineLearningSurgical2021}.
Current state-of-the-art methods typically combine convolutional backbones with LSTM or attention-based temporal accumulators. 
These methods are particularly well suited for longer videos, as frequently seen in the surgical domain, where acquisitions span many hours~\cite{funkeUsing3DConvolutional2019,czempielTeCNOSurgicalPhase2020,czempielOperAAttentionRegularizedTransformers2021}.

\begin{figure}
  \centering
    \subfloat{\includegraphics[width=0.99\textwidth]{"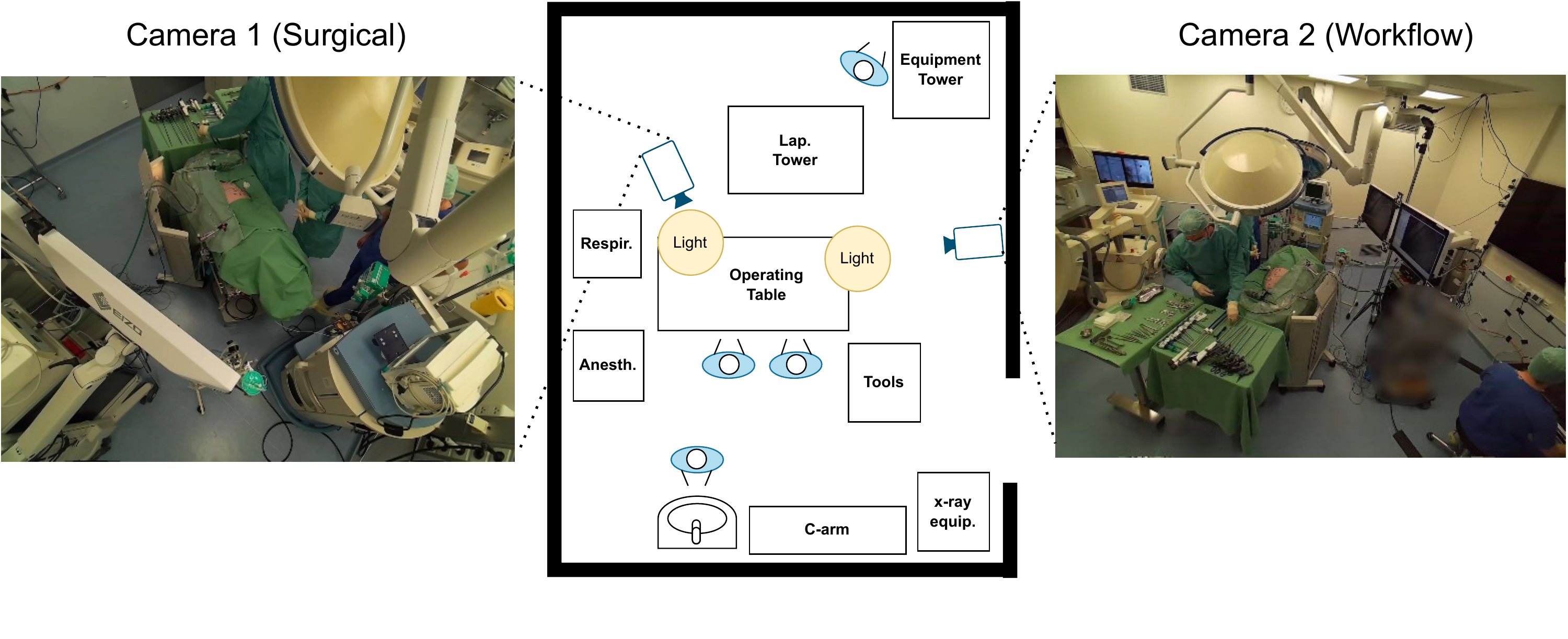"}}%
  \caption{\textbf{Multi-view OR Dataset}. Camera 1 (Surgical Camera) is positioned above the laparoscopic tower facing downwards, approximately 1.5 m from the operating table. Camera 2 (Workflow Camera) provides a wide angle view of the procedure, positioned about 2.5 m away from the operating table. Both cameras are ceiling mounted, providing two different perspectives of the scene.}
  \label{fig:or_cameras}
\end{figure}

Despite their prevalence in the computer vision community~\cite{feichtenhoferX3DExpandingArchitectures2020}, the adoption of clip models in the medical domain has been limited.
This is partially attributable to the length of distinguishable actions in surgical videos.
In the medical domain, events of interest may appear indistinguishable on such a short timescale, requiring the aggregation of additional temporal information. 
Clip-level architectures, however, remain relevant even for longer videos, as they can be similarly combined with LSTMs to aggregate both short and long-term image features~\cite{schmidtMultiviewSurgicalVideo2021}. 
Extracting clip features can provide performance benefits as they reduce dimensionality compared to image level features \cite{sharghiAutomaticOperatingRoom2020}. 
Furthermore, they can extract information from short video segments where training complex temporal models may be infeasible. 

\textbf{Multi-sensor Fusion.}
Depth sensors have been adopted extensively in the medical domain due to their privacy-preserving qualities~\cite{banerjeeMonitoringPatientsHospital2014,magiActivityMonitoringICU2020}.
However, it is not always clear how well complex actions can be represented solely from depth images.
It has been suggested that surgical gesture recognition can be learned from optical flow data alone, highlighting the importance of motion cues for action recognition~\cite{sarikayaSurgicalGestureRecognition2020}.
Combining modalities from different sensors continues to be an active topic in the medical domain~\cite{huaulmePEgTRAnsferWorkflow2022}, as data stemming from various sources become ubiquitous in the OR.

Fusion strategies for RGB-D data are being explored at length for many
applications such as depth estimation~\cite{jung2021wild}, 6DoF pose estimation~\cite{saadiOptimizingRGBDFusion2021,busam2018markerless}, object classification ~\cite{shaoPerformanceEvaluationDeep2017}, and also video action recognition \cite{shaikh2021rgb}.
Due to significant differences between these modalities, late fusion strategies are frequently used to combine 3D geometric information with texture from RGB images.
However, early fusion via concatenation of RGB-D image channels has been shown to be beneficial in combination with specific network architectures~\cite{zhaoSingleStreamNetwork2020}.

We tackle surgical action recognition from multi-modal image clips. In our problem setting, a clip consists of a short image sequence from an individual modality (RGB, IR, depth) or multiple modalities in combination (see Fig.~\ref{fig:pipeline}). Our experiments are carried out on a series of novel experimental multi-view OR acquisitions, well suited to illuminate how current state-of-the-art video action recognition methods cope with different sensor inputs in surgical environments.

\section{Dataset}
Our dataset consists of 16 laparoscopic surgeries on an animal model (swine), of experienced surgeons testing a novel augmented reality laparoscope.
We captured the RGB-D video streams using two ceiling-mounted Kinect-Azure \cite{azuresdk} cameras (cf. Fig.~\ref{fig:or_cameras}). Eight medically relevant workflow phases were identified together with expert clinicians, and annotated for subsequent action classification. The resulting action segments account for over 24 hours of video, containing roughly 2.5 million frames per view.

\begin{itemize}
\item Room Disinfection (\textbf{RDS}). This phase is distinguished mostly by the lack of patient presence in the operating room. Nurses, anesthetists, and research scientists are frequently present performing auxiliary tasks.
\item Rolling in (\textbf{PIN}) and rolling out (\textbf{POUT}) of the patient before / after surgery.
The swine is disconnected from cables and infusion systems for patient-out, moved onto a portable table, draped, and rolled out of the room by two to four staff members.
Patient-in is characterized by the reverse sequence of these actions.
\item Patient preparation (\textbf{PREP}). The surgical team implants an artificial tumor by laparotomy (cutting through the abdominal wall) of the swine.
At the same time, the right jugular vein and carotid artery are prepared for the insertion of a central venous catheter and an arterial sluice by the veterinary staff. 
These actions occurred asynchronously for some trials.
\item Patient monitoring (\textbf{MON}). Patient is present in the scene but no surgical or preparatory procedures are being performed. An anesthesist monitors the patient and equipment.
\item The surgeon inserts an arterial catheter through the arterial sluice into the abdominal aorta, then conducts an x-ray guided catheter angiography (\textbf{LAC}) with the use of a C-arm.
\item A laparascopic kidney, rectal, or pancreas resection is performed (\textbf{SURG}). These are treated identically for our task of workflow phase recognition.
\item After the laparoscopic procedure, a conventional laparotomy, and tumor resection are carried out, followed by a subsequent suturing (\textbf{TRS}).
\end{itemize}

\begin{figure}
  \centering
    \includegraphics[width=0.81\textwidth]{"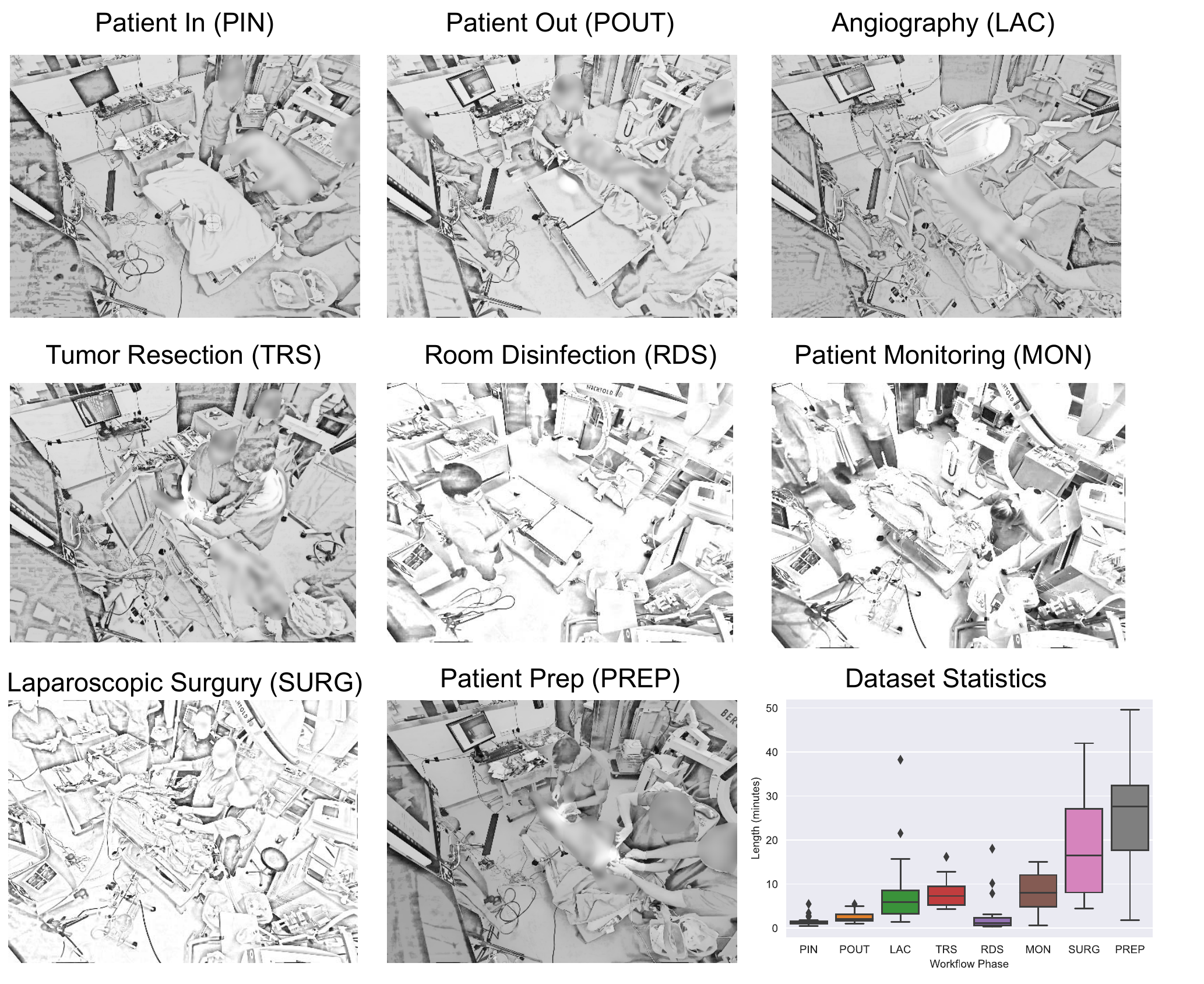"}%
  \caption{\textbf{Surgical Phase Acquisitions}. Artistic renditions (for privacy reasons) of characteristic frames
  for each surgical action class in our dataset, along with box plots describing the distributions of their length. Inter- and intra- phase lengths vary significantly, containing several outliers. Images are from the perspective of the Surgical Camera (Camera 1).}
  \label{fig:modalities}%
\end{figure}

Due to the experimental nature of these procedures, the eight phases consist of disjoint action sections varying significantly in defining characteristics and length (see Fig.~\ref{fig:modalities}).
For instance, patient-in averages slightly over 30 seconds in length, making the long-term accumulation of temporal features challenging.

\section{Methodology}
To better understand how different input modalities impact surgical action recognition, we study two commonly used backbone architectures: ResNet50~\cite{heDeepResidualLearning2015}, and its temporal clip extension X3D~\cite{feichtenhoferX3DExpandingArchitectures2020}. 

\textbf{Network Architecture.}
Convolutional architectures such as ResNet-50 have been extensively used for phase segmentation of endoscopic videos.
They serve as feature extraction backbones for many state-of-the-art recognition architectures~\cite{jinMultiTaskRecurrentConvolutional2019,czempielOperAAttentionRegularizedTransformers2021,yuanSurgicalWorkflowAnticipation2021}.
Our baseline consists of a ResNet-50 pre-trained on ImageNet without temporal modeling.

We further adopt the X3D-M~\cite{feichtenhoferX3DExpandingArchitectures2020}
architecture for clip-level classification, due to its advantageous efficiency-performance trade-off for action recognition. 
Following recent trends in video classification for smaller datasets, we use an X3D model pre-trained on Kinetics-400~\cite{bertasiusSpaceTimeAttentionAll2021} for all experiments.

    Clip-based video architectures, such as X3D and I3D, incorporate temporal information differently from traditional Spatio-temporal models such as HMMs, RNNs, or LSTMs, which typically aggregate temporal information from pre-extracted image features~\cite{carreiraQuoVadisAction2017}. Given an image $x \in \mathbf{R}^{w \times h}$, we sample $t$ images from a temporal window $T$ with a constant temporal stride $\tau$. X3D and I3D extract features from all frames jointly by passing the stacked image clips $x_t \in \mathbf{R}^{w \times h \times t}$ directly into a 3D convolutional backbone. These architectures induce less bias towards image-level feature extraction than traditional Spatio-temporal models and allow fine-grained control on kernel inflation across the temporal dimensions~\cite{feichtenhoferX3DExpandingArchitectures2020}. The resulting architectures are typically lightweight, and extracted features can still be paired with long-term temporal accumulators, as has been demonstrated for surgical action recognition~\cite{sharghiAutomaticOperatingRoom2020}. Furthermore, clip-based architectures can be trained to produce a prediction for a clip's last frame, making them online-capable. They operate in real-time --- we produced predictions from RGB clips at over 90 frames-per-second on our architecture.

    In our experiments, clips are randomly sampled with a constant temporal stride $\tau$ of four fps.
    A clip length $t=16$ frames therefore results in a four second-long temporal window $T$.

\textbf{Model Validation.}
We present a robust evaluation inspired by \cite{nwoye2022data} to rigorously quantify the differences between modalities across model architectures, and camera views, eliminating potential bias caused by evaluating on a single small test set. The 16 trials were randomly split into four non-overlapping test folds. For each test fold, two of the remaining three folds are used for training, while one is used for model validation. Each model/modality/view combination is trained across all four splits.

\textbf{Image Modalities and Fusion.}
We proceed by comparing the input modalities $M$ captured by the Kinect cameras individually and in two combinations of particular interest (see Fig.~\ref{fig:pipeline}). RGB and depth images are fused by concatenation to generate a four-channel image according to previous early fusion approaches~\cite{zhaoSingleStreamNetwork2020}.
Furthermore, we attempt to reproduce the depth-ir fusion from~\cite{sharghiAutomaticOperatingRoom2020,schmidtMultiviewSurgicalVideo2021}, by color-coding the depth data and alpha-blending them with the raw IR images.
All images are augmented using a small random affine transformation followed by cropping the 224x224 pixel center.
Minor brightness and contrast adjustments are applied to the RGB images.
We add a small amount of Gaussian noise proportional to the magnitude of each ray to the depth images.
These augmentations are applied to the relevant channels separately for early fusion before concatenation.

We note that RGB pre-trained networks cannot extract features from depth images adequately.
This is likely due to the difference between the information encoded in these modalities~\cite{zhaoSingleStreamNetwork2020} -- the texture represented in RGB images differs significantly from the geometry encoded in depth images. 
Therefore, during fine-tuning, we un-freeze only the first two ResNet encoder blocks corresponding to the feature extraction components of the network in addition to the prediction head.
This allows the networks to adapt to each input modality while preserving the rich action recognition embedding learned from Kinetics-400.
We followed this transfer strategy for all input modalities to provide an objective comparison.

We additionally construct an approach for late fusion of RGB and depth to combine an arbitrary number of pre-trained modality-specific models. The feature vector predictions $f(M) \in \mathbb{R}^{2048}$ from each network are concatenated and passed through a series of three fully connected layers, which culminate in a prediction $p$. For both early and late fusion strategies of RGB and depth, the two modalities were first aligned by warping depth into the perspective of the color camera using the Kinect Azure SDK \cite{azuresdk}.

\textbf{Training Details.}
Hyperparameters for all augmentations are tuned using hyperband search \cite{li2017hyperband} and fixed for all model and modality types. All optimization hyperparameters are tuned separately for the resnet and X3D models but preserved across modalities and views.
We then train the final model for each modality for 300 epochs with the previously described transfer scheme.
During each epoch, we sample an equal number of frames or video clips from each video in the training set in order to combat class imbalance due to heterogeneous video lengths.
We then fine-tune the entire network for an additional 150 epochs by un-freezing all network layers.
Models are trained with momentum SGD, a cosine annealing weight decay, with a weight regularization of $10^{-5}$. Fine-tuning is performed at a constant learning rate of $10^{-4}$.
A cross-entropy loss was used to learn the classification task. All models were trained using Pytorch v1.12 \cite{pytorch2019}, on a single Nvidia A40 GPU. 

To quantify the prediction performance of each network, we measure the mean average precision (mAP)~\cite{idrees2017thumos,nwoye2022data} and accuracy (acc). Both are balanced across classes due to our sampling strategy.

\begin{table}
\setlength{\tabcolsep}{7pt}
\renewcommand{\arraystretch}{1.2}
\centering
\caption{Comparison of individual modalities, as well as early (ef) and late fusion (lf) strategies for two OR views. Accuracy (acc) and mean average precision (mAP) are reported for each experiment, in percentage, ± standard deviation over four validation folds. Values in bold indicate the best-performing models over single and fusion modalities, respectively.}
\label{tab1}
\resizebox{0.99\textwidth}{!}{
\begin{tabular}{l  c  c  c  c | c c c c}
\toprule
  & \multicolumn{4}{c|}{Camera 01 (Surgical)} & \multicolumn{4}{c}{Camera 02 (Workflow)} \\
  & \multicolumn{2}{c}{ResNet} & \multicolumn{2}{c|}{X3D} & \multicolumn{2}{c}{ResNet} & \multicolumn{2}{c}{X3D} \\
\midrule
  Modality & acc. & mAP & acc. & mAP & acc. & mAP & acc. & mAP\\
\midrule
  RGB        & 72.9 ± 2.7 & 84.5 ± 1.6 & \textbf{82.4 ± 4.5} & \textbf{90.7 ± 2.6} 
             & 75.4 ± 1.6 & 85.8 ± 2.9 & \textbf{82.2 ± 7.0} & \textbf{89.6 ± 6.6} \\
  Depth      & 68.0 ± 4.7 & 78.1 ± 3.4  & 78.0 ± 3.2 & 83.9 ± 2.8
             & 60.6 ± 5.9 & 67.6 ± 4.8 & 69.8 ± 9.1 & 75.0 ± 4.2 \\ 
  IR         & 62.7 ± 9.7 & 72.2 ± 8.9 & 78.1 ± 4.3 & 83.4 ± 1.7
             & 61.3 ± 0.3 & 70.3 ± 1.2 & 75.5 ± 5.1 & 81.9 ± 3.2 \\ 
\midrule
  IR+Depth${}^{\text{ef}}$ 
             & 67.8 ± 4.7 & 73.7 ± 4.0 & 73.4 ± 5.0 & 80.3 ± 3.7
             & 63.4 ± 3.4 & 68.4 ± 4.3 & 67.0 ± 8.7 & 69.4 ± 6.7 \\              
  RGB+Depth${}^{\text{ef}}$
             & 73.5 ± 3.6 & 86.1 ± 1.8 & 70.76 ± 4.3 & 80.4 ± 6.3
             & 68.2 ± 7.1 & 79.5 ± 6.2 & 75.8 ± 7.4 & 84.8 ± 4.7 \\ 
  RGB+Depth${}^{\text{lf}}$
             & 76.7 ± 4.3 & 86.7 ± 3.2 & \textbf{85.0 ± 1.6} & \textbf{92.0 ± 2.1}
             & 75.4 ± 5.4 & 83.5 ± 4.2 & \textbf{84.8 ± 5.4} & \textbf{89.4 ± 4.6}\\ 
\bottomrule
\end{tabular}
}
\end{table}

\section{Evaluation}
\textbf{Results.} Several notable trends become apparent from our modality study (Table \ref{tab1}). 

Incorporating temporal information into the backbone using X3D increases accuracy and mAP across all modalities and views by up to $10\%$, even for fusion architectures. Furthermore, the disparities between input modalities are generally consistent for both architectures and camera views.

RGB consistently outperforms depth and IR for both ResNet and X3D architectures.
Depth and IR display a comparable performance across both architectures and views.
In our experiments, the fusion of depth and IR via alpha-blending did not provide improvements over either individual modality.
In contrast, the combination of RGB and depth resulted in varying performance.
Accuracy and mAP improve significantly for the late fusion of RGB and depth data, with X3D late fusion exhibiting the highest mAP for both the surgical camera (92.0 ± 2.1) and workflow camera (89.4 ± 4.6). 
However, RGB and depth early fusion seems to suffer compared to RGB alone.

In summary, late fusion offers the most considerable improvements in accuracy and mAP, while RGB performs significantly better than other individual modalities. These results are consistent across model architectures and camera views.

We additionally compare the differences in complexity between the frame and clip-based methods with the I3D~\cite{carreiraQuoVadisAction2017} backbone commonly used for surgical action recognition. While the X3D architecture requires fewer floating point operations (FLOPs) than I3D for a forward pass, the combined loading and processing of two modalities in the late fusion approach incurs a considerable additional training cost (Table \ref{tab2}).

\begin{figure}
  \centering
    \includegraphics[width=0.99\textwidth]{"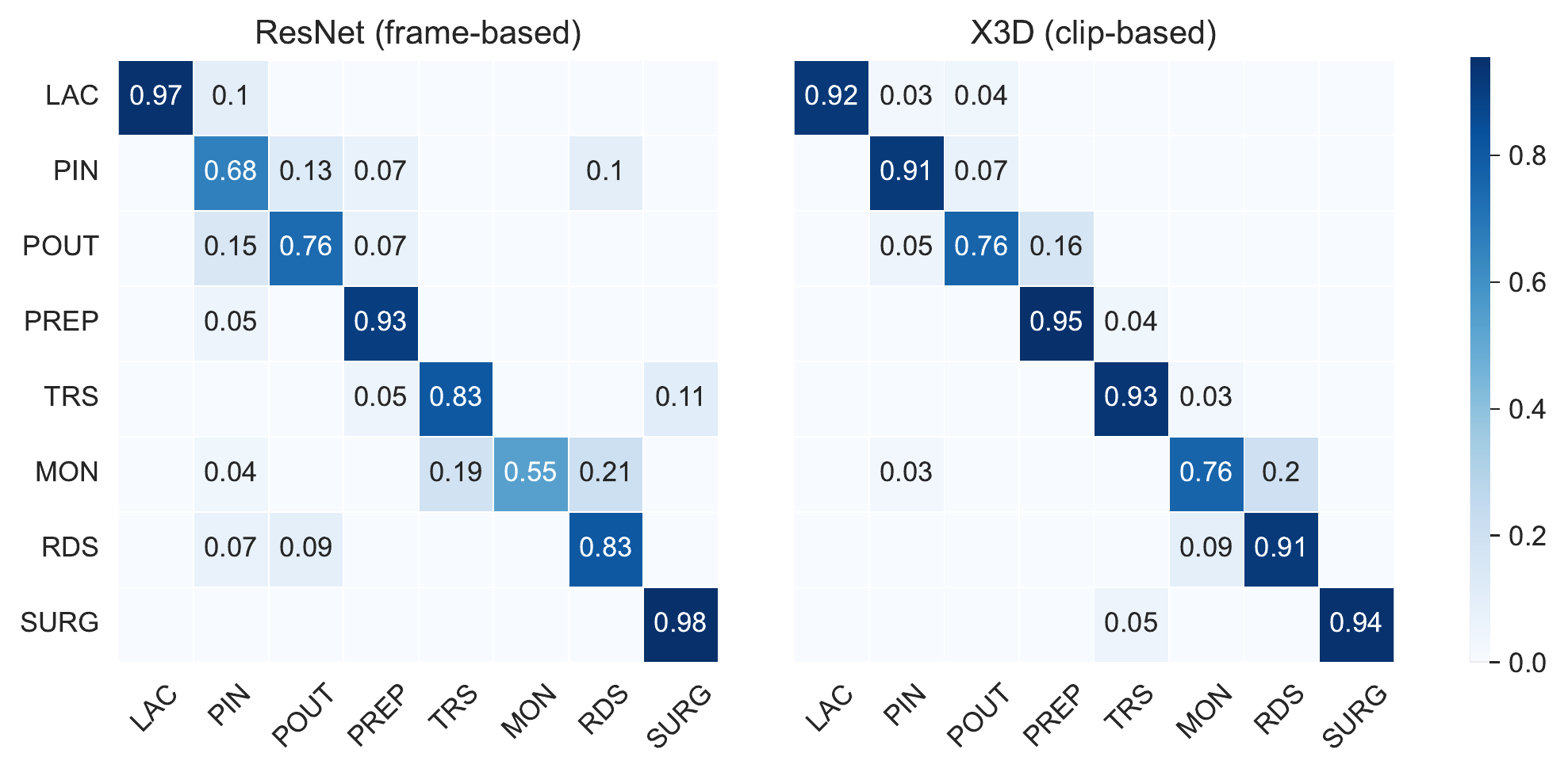"}%
  \caption{\textbf{Confusion Matrix of the best performing modalities - RGB-D late fusion, Workflow Camera, comparing ResNet and X3D.}
  Several temporally ambiguous classes contribute to the differences between frame and clip class level classification accuracy. The clip level model better distinguishes phases such as PIN, POUT, and MON. Values below $0.03$ are omitted for clarity.}
  \label{fig:cm_rgbd}%
\end{figure}

\begin{figure}
  \centering
    \includegraphics[width=0.99\textwidth]{"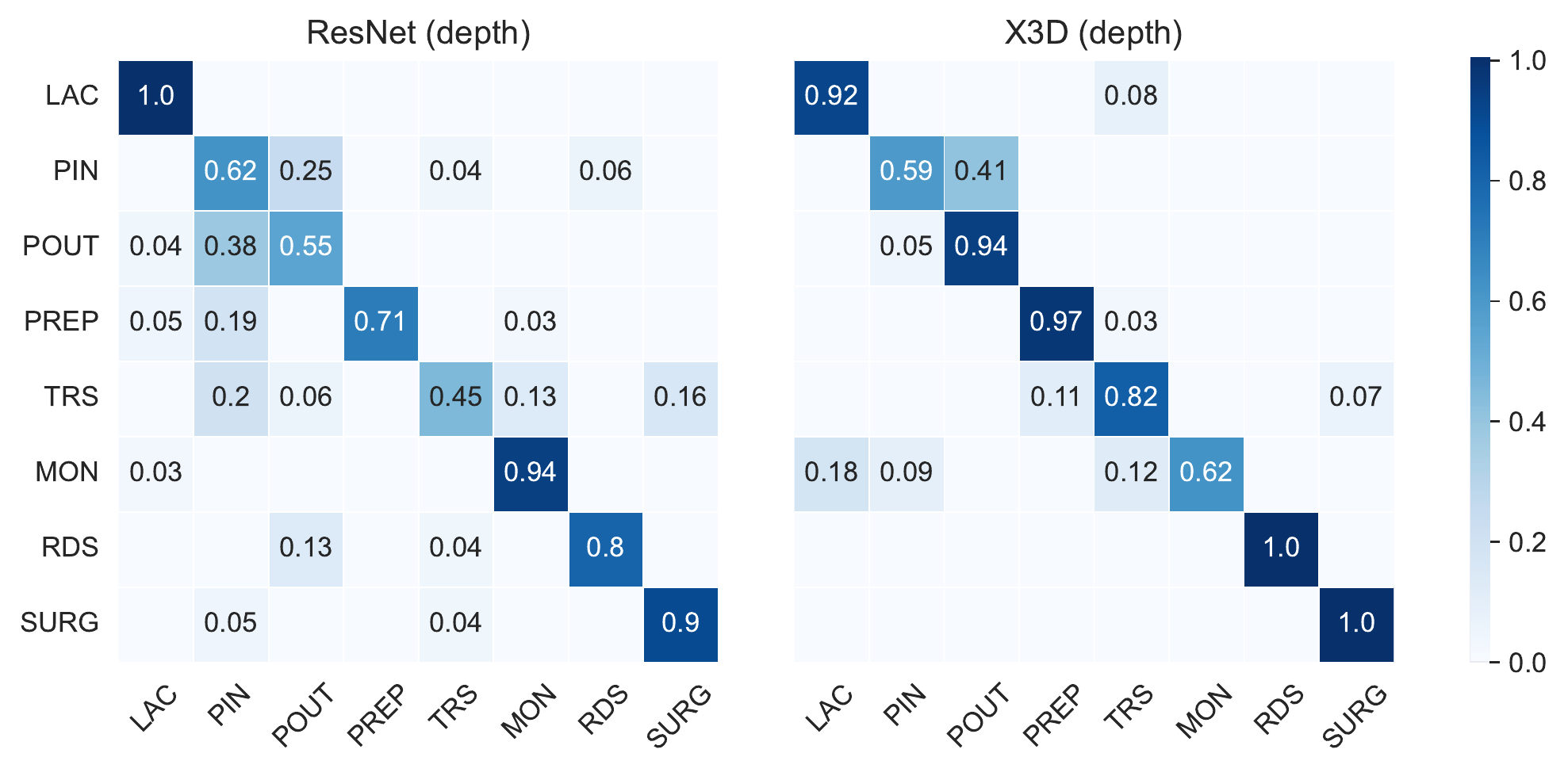"}%
  \caption{\textbf{Confusion Matrix depth-only, Surgical Camera.}
 Motion cues derived from temporal depth sequences lead to large overall performance gains. Depth-only performs particularly poorly on distinguishing temporally ambiguous classes such as patient in (PIN) and out (POUT). Values below $0.03$ are omitted for clarity.}
  \label{fig:cm_depth}%
\end{figure}

\textbf{Discussion.} 
Incorporating temporal features is critical to developing highly performant feature extraction backbones. Surgical workflow phases frequently exhibit high degrees of temporal ambiguity. A notable example from our evaluation can be seen by comparing confusion matrices across model architectures (Fig. \ref{fig:cm_rgbd}), where the frame level model fails to distinguish patient in (PIN) and patient out (POUT) adequately. These phases are difficult to segregate unless the direction of motion is clear. A similar pattern can be observed when comparing the confusion matrices of depth-only across the image and clip-based backbones (Figure \ref{fig:cm_depth}). The clip-based model obtains significantly higher accuracy across these classes. Furthermore, a high degree of ambiguity appears to exist between the phases MON, TRS, and RDS at a frame level.

Our results suggest that RGB is critical for the task of surgical workflow recognition. The information-rich color space contains important visual cues our RGB models learn to utilize. 
Even a combination of IR and depth provide meager performance compared to using RGB alone. This is evident for both the X3D backbone, but also the backbone used in~\cite{sharghiAutomaticOperatingRoom2020,schmidtMultiviewSurgicalVideo2021} --- we see a significant performance drop compared to even a frame-level resenet model with only RGB as input (Table \ref{tab2}). Notably, depth data also contains a significant amount of information for this task in our environment which convolutional architectures learn to utilize. 
For the surgical camera, depth results fall less than $5\%$ short of RGB with respect to single modalities. However, the discrepancy is much larger for the workflow camera (Fig. \ref{fig:cm_depth}), with more than $10\%$ difference in accuracy and mAP. 
This could be due to the limited active range of the ToF sensor, indicating camera placement is crucial when relying on depth alone. 
In cases where RGB data cannot be used due to privacy concerns, this should be taken into consideration. 
Cameras relying on depth or IR only could fail to adequately distinguish between phases at a larger distance. 
In contrast, the RGB based architectures perform similarly across views in terms of absolute metrics, despite significantly different viewing angles and distance from the center of the OR.

For both cameras, depth is vital in improving performance over RGB alone.
This is especially evident for late fusion models, which outperform all other models for both camera positions. Geometric structures portrayed in depth are robust against lighting and changes in texture, providing valuable additional information for processing complex scenes~\cite{zhaoSingleStreamNetwork2020}. The importance of depth for surgical action classification is evident from our fusion results.

Late fusion performs best across all our experiments, suggesting that independent feature extraction before fusion may be beneficial for surgical workflow analysis. Notably, RGB benefits from these additional cues for distinguishing behavior in a scene precisely due to the inherent differences between these two data types. Furthermore, the modalities may be different enough to warrant using separate or even specialized encoders or network architectures~\cite{jung2021wild,qi2017pointnet,zhaoSingleStreamNetwork2020} to address the inherent geometric properties of depth data. Compared to the computationally expensive late fusion architectures, this could provide an advantageous performance/accuracy trade-off. We leave the comparison of additional slow fusion methods open for future exploration.

\begin{table}[!h]
\setlength{\tabcolsep}{7pt}
\renewcommand{\arraystretch}{1.2}
\centering
\caption{\textbf{Choose your backbone wisely.} We summarize the performance of different common backbone architectures used for surgical phase prediction trained on our in-house dataset. Accuracy and mAP are listed, as well as GFLOPs (billion floating point operations) and average training time on a single Nvidia A40 GPU. Note that X3D and I3D are clip-based architectures, while ResNet is image based. We only train the model backbones in order to compare fairly across methods, i.e. without additional long-term memory components (RNN, LSTM, etc.).}
\label{tab2}
\resizebox{0.99\textwidth}{!}{
\begin{tabular}{c c c c c c c c}
\toprule
  Backbone & Modalities & Camera 01 Acc. & Camera 01 mAP & Camera 02 Acc. & Camera 02 mAP & GFLOPs\footnote{derived from \cite{fan2021pytorchvideo}} & Training Time (hrs) \\
\midrule
Proposed (X3D) & RGB+Depth lf & 85.0 ± 1.6 & 92.0 ± 2.1 & 84.8 ± 5.4 & 89.4 ± 4.6 & 13.4 & 19.93 \\
I3D \footnote{\cite{sharghiAutomaticOperatingRoom2020,schmidtMultiviewSurgicalVideo2021}} & IR+Depth ef & 75.0 ± 4.6 & 82.4 ± 4.2 & 66.6 ± 5.7 & 67.8 ± 3.6 & 37.5 & 7.51 \\
ResNet \footnote{\cite{jin2017sv,jinMultiTaskRecurrentConvolutional2019, czempielTeCNOSurgicalPhase2020,nwoyeRendezvousAttentionMechanisms2021, czempielOperAAttentionRegularizedTransformers2021}} (without LSTM) & RGB &  72.9 ± 2.7 & 84.5 ± 1.6 & 75.4 ± 1.6 & 85.8 ± 7.0 & 4.0 & 0.64 \\
\bottomrule
\end{tabular}
}
\end{table}


In summary, these findings highlight the importance of using RGB for scene-level recognition tasks in our OR settings. 
Incorporating an additional depth stream to RGB provided significant improvements over RGB only for our early fusion approach. 
These differences hold across two commonly used model architectures and camera views. 
It has been shown that a more capable visual backbone directly correlates with better prediction results for temporal models~\cite{czempielTeCNOSurgicalPhase2020}, making modality fusion an indispensable step toward more accurate surgical action classification.

\section{Conclusion}

In this work, we systematically demonstrate that multi-modal data fusion improves automated surgical action recognition, an essential task for context-aware systems in the OR. We methodically analyze how modality impacts surgical action recognition on a multi-view surgical OR dataset. Our RGB-D late fusion method consistently performs better than either individual modalities across model architectures and camera views. Incorporating temporal information in the form of image clips significantly provides large accuracy gains across all modalities. Future OR layouts should contextualize these differences in sensor modality performance for the specific recognition task at hand. We are confident that these findings pave the way for further exploration into how sensor modalities can coordinate under the adverse lighting conditions or occlusions ever-present in surgical operating environments.

\hfill \break
\textbf{Ethical.}
All procedures were carried out in strict accordance with recommendations and guidance for the care and use of laboratory animals of the National Institutes of Health, which received full approval by the local Ethical Committee on Animal Experimentation by Germany and Ludwig Maximilian University (LMU) ROB-55.2-2532.Vet\_02-20-212.

\hfill \break
\textbf{Acknowledgements.}
This work was funded by the German Federal Ministry of Education and Research (BMBF), No.: 16SV8088 and 13GW0236B.

\bibliographystyle{tfcse}
\bibliography{workflow_phase}

\end{document}